\documentclass{article}

\usepackage{arxiv}

\usepackage{amsmath,amsthm,amssymb,amscd}
\usepackage[all,cmtip]{xy}
\usepackage{diagxy}
\usepackage[utf8]{inputenc} 
\usepackage[T1]{fontenc}    
\usepackage{hyperref}       
\usepackage{url}            
\usepackage{booktabs}       
\usepackage{amsfonts}       
\usepackage{nicefrac}       
\usepackage{microtype}      
\usepackage{lipsum}
\usepackage{mathrsfs}
\usepackage{fancyhdr}
\usepackage[dvips]{color}
\usepackage{color}
\usepackage{subfig}
\usepackage{graphicx}

\newtheorem{exam.}[subsubsection]{Example}
\newtheorem*{def.}{Definition}

\newtheorem{lemma}[subsubsection]{Lemma}
\newtheorem{prop.}[subsubsection]{Proposition}

\newtheorem{theorem}[subsubsection]{Theorem}
\newtheorem*{remark}{Remark}

\newcommand{\shF}{\mathscr{F}}

\newcommand{\shG}{\mathscr{G}}

\title{A Sheaf and Topology Approach to \\ Generating Local Branch Numbers in Digital Images}
\author{
  Chuan-Shen Hu \\ 
  Department of Mathematics\\
  National Taiwan Normal University\\
  Taipei, Taiwan \\
  \texttt{peterbill26@hotmail.com}
  \AND
  Yu-Min Chung \\ 
  Department of Mathematics and Statistics\\
  University of North Carolina at Greensboro\\
  Greensboro, USA  \\
  \texttt{y\_chung2@uncg.edu} \\
}

\begin{document}
\maketitle

\begin{abstract}
This paper concerns a theoretical approach that combines topological data analysis (TDA) and sheaf theory. Topological data analysis, a rising field in mathematics and computer science, concerns the shape of the data and has been proven effective in many scientific disciplines. Sheaf theory, a mathematics subject in algebraic geometry, provides a framework for describing the local consistency in geometric objects.  Persistent homology (PH) is one of the main driving forces in TDA, and the idea is to track changes of geometric objects at different scales.  The persistence diagram (PD) summarizes the information of PH in the form of a multi-set. While PD provides useful information about the underlying objects, it lacks fine relations about the local consistency of specific pairs of generators in PD, such as the merging relation between two connected components in the PH. The sheaf structure provides a novel point of view for describing the merging relation of local objects in PH.  It is the goal of this paper to establish a theoretic framework that utilizes the sheaf theory to uncover finer information from the PH.  We also show that the proposed theory can be applied to identify the branch numbers of local objects in digital images.
\end{abstract}
\section{Introduction}
\begin{figure*}
\begin{center}
\subfloat[Two filtrations of binary images]{\includegraphics[scale=0.60]{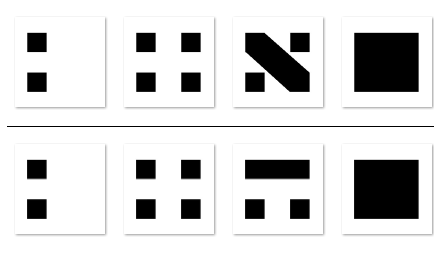}}~
\subfloat[The persistence diagram $\mathcal{P}_0$]{\includegraphics[scale=0.30]{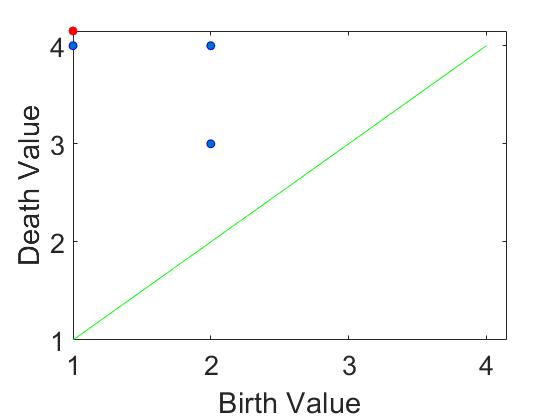}}
\end{center}
\centering
\caption{An example of two filtrations that share the same persistence diagram.}
\label{FIG : Example of two filtations share the same PD}
\end{figure*}

Topological data analysis (TDA) is a branch of applied mathematics that aims to quantify topological characteristics, specially the $q$-dimensional Betti numbers denoted by $\beta_q$.  For instance, $\beta_0$, $\beta_1$, and $\beta_2$ represent number of components, holes, or voids, respectively. \textit{Persistent homology} (PH), one of main tools in TDA, tracks changes of filtered topological spaces induced by the original objects~\cite{PH2002, PH2005, TDA2009, PH2014, Gudhi_Lib_2014, Bauer2017PhatP}. Figure~\ref{FIG : Example of two filtations share the same PD} (a) show two examples of filtered spaces (also known as  \textit{filtration}) of black regions. The $q$-dimensional PH of a \textit{filtration} of topological spaces is a sequence of vector spaces that are connected by linear transformations\footnote{In algebraic topology, one may also consider PH of modules and module homomorphisms.}. The non-negative integer $q$ denotes the dimension of objects which are captured by the PH. For instance, the PH of $q = 0$ captures the changes of connected components and $q = 1$ for holes in a filtration.

\begin{figure}
\begin{center}
\subfloat[$g_1$]{\includegraphics[scale=0.35]{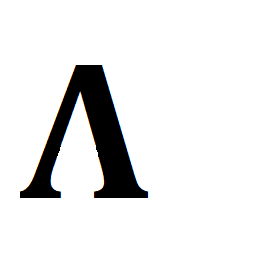}}
\subfloat[$g_2$]{\includegraphics[scale=0.35]{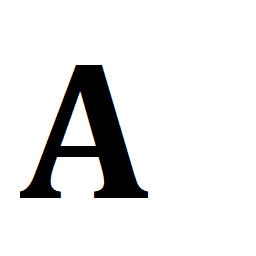}}
\subfloat[$g_3$]{\includegraphics[scale=0.35]{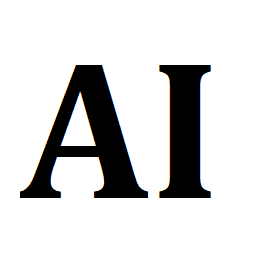}}
\subfloat[$g_4$]{\includegraphics[scale=0.35]{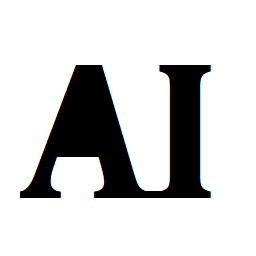}}\\
\subfloat[$\mathcal{P}_0$]{\includegraphics[scale=0.35]{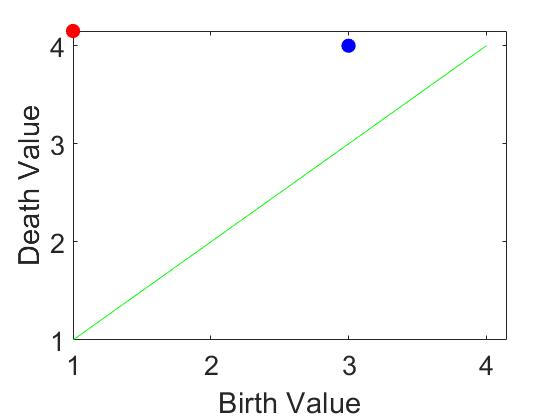}}~
\subfloat[$\mathcal{P}_1$]{\includegraphics[scale=0.35]{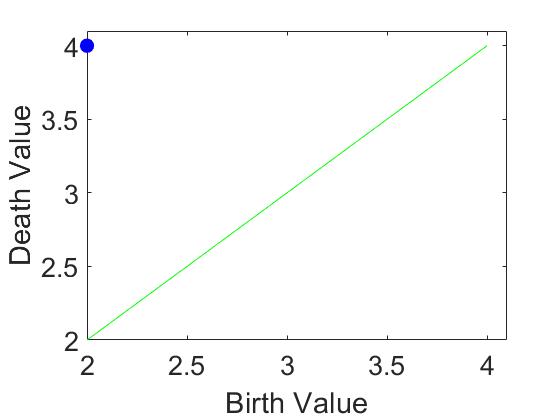}}~
\end{center}
\centering
\caption{(a) $\sim$ (d) is a filtration of black pixels. (c) Binary image ``AI'', which has $\beta_0 = 2$ (connected components) and $\beta_1 = 1$ ($1$-dimensional holes). (e) and (f) are persistence diagrams $\mathcal{P}_0 = \{ (1, \infty), (3, 4) \}$, $\mathcal{P}_1 = \{ (2, 4) \}$ in dimension $0$ and $1$ of the filtration respectively.}
\label{FIG : Example of image filtration}
\end{figure}

Persistence barcodes (or simply \textit{barcode}) are a typical way to summarize the information of PH.  A barcode of a $q$-dimensional object/generator in PH is a pair of numbers $(b,d)$ where the object/generator is born at the value $b$, and dies at the value of $d$.  One often refers $b$, and $d$ to the \textit{birth} and \textit{death}, respectively, values.  For instance, in Figure~\ref{FIG : Example of image filtration}, an $1$-dimensional hole of the character ``A" was born at $g_2$ and disappeared at $g_4$, and hence the hole has the barcode $(2,4)$. Barcodes in PH can be defined rigorously as the algebraic structure of PH~\cite{PH2002, PH2005, TDA2009, PH2014} (cf. Lemma \ref{App Lemma}). The collection of all barcodes of $q$-dimensional objects is called the \textit{persistence diagram} (PD)~\cite{PH2002, PH2005, TDA2009, PH2014}. PD plays an essential role in the application, and has widely applied and studied in computer vision~\cite{8953990,ICMV2020TDACv,DBLP:journals/corr/abs-1906-01769}, machine learning~\cite{JMLR:v16:bubenik15a, JMLR:v18:16-337, JMLR:v18:17-317, pun2018persistenthomologybased}, image/signal processing~\cite{ PHandNLPZhu2013, Boche2013}, medical science~\cite{MATE20141180, Kim2012, Kim2015, bendich2014persistent, Biscio2016TheAP}, and physics~\cite{spitz2020finding, Anand2020}.

Although PD contains information about the changes of local objects, the reduction algorithm for obtaining Smith normal forms~\cite{EdelsbrunnerHarerbook2010} of the connecting linear transformations would omit the merging relations of elements in PH. In other words, the PD is not enough for researchers who are interested in the merging behaviors of certain local objects. For example, top and bottom panels of Figure~\ref{FIG : Example of two filtations share the same PD} (a) show two different filtrations of binary images, but they share the same persistence diagram $\mathcal{P}_0$ as shown in Figure~\ref{FIG : Example of two filtations share the same PD} (b). The merging relation between connected components at the third image is different: one connects diagonally while the other connects horizontally.

There are some work related to the behaviors of local objects in PH, such as Mapper~\cite{Kraft900997,Singh2007TopologicalMF} and local (co)homology~\cite{Local(co)PH, Local(co)PHFasy2016}. In particular, Vandaele \textit{et al.}~\cite{Vandaele_Local_TDA} investigated the local Betti numbers $(\beta_0, \beta_1)$ of point clouds by computing local Vietoris-Rips complexes and their PHs, where the local $\beta_0$ and $\beta_1$ corresponds to the branch numbers and holes of local objects. For a point cloud, the local Betti pairs provide a heat map of branch numbers and loop structures. Comparing to the global Betti numbers, it is an additional information for the point cloud. The proposed work bases on a similar idea and provides a sheaf theoretical approach to describe the local behaviors of a geometric object.  

Historically, sheaves were first developed as a tool for researching the nerve theory and fixed-point theorems~\cite{AlexandroffNerve,levy1952foundations,bredon1997sheaf}, while the recent trend is to study algebraic geometry~\cite{Hartshorne}.  Sheaf theory and algebraic geometry provide fruitful results and tools in analyzing local/global properties of geometric objects, while the research of the combination of sheaves and TDA is still in its infancy.   A sheaf $\shF$ over a topological space $X$ is a rule which assigns each open subset $U$ of $X$ to an algebraic object $\shF(U)$. Except objects, a sheaf $\shF$ also assigns each pair $V \subseteq U$ of open sets to a homomorphism $\rho_{UV} : \shF(U) \rightarrow \shF(V)$ as an connection between two algebraic objects. The assignments of $U$ and $V \subseteq U$ are analogous to the space of functions on $U$ and restriction of functions on $V$. 

Recently, the combination of the sheaf theory and TDA has been found its potential in modeling and analyzing real data~\cite{ghrist2014elementary, RobinsonNyquist, RobinsonHunting, robinson2014topological, berkouk:tel-02976867, hansen2020opinion, Curry2015}. For example, Robinson~\cite{RobinsonNyquist} proved the Nyquist sampling theorem by computing cellular sheaf cohomologies on abstract simplicial complexes of real signals and their samplings. In the work, the concept of \textit{global and local sections} on posets equipped with the \textit{Alexandrov topology}~\cite{Ale37, Ale47} is crucial for constructing the sheaf structures on digital/analog signals.  Our work is also motivated by the \textit{multi-parameter persistent homology}~\cite{CarlssonSinghZomorodian, SCARAMUCCIA2020101623, harrington2019stratifying} and zigzag homology~\cite{carlsson2008zigzag,Carlsson09zigzagpersistent, carlsson2019persistent, carlsson2019}.  The extension of barcodes to multi-parameter persistent homology does not exist, and is an active research area in TDA.  As discussed in \cite{Curry2015}, sheaf theory may be an appropriate tool towards that goals.  The cellular sheaves considered in the paper can be viewed as a form of zigzag homology and multi-parameter persistent homology.  The proposed work aims to capture local sections of certain local objects and their geometric meaning in digital images.    

Our main contributions can be summarized in the followings. 
\begin{itemize}
    \item We propose the \textit{coincidence} of pairs in the filtration of binary images. The coincidence can be identified as global/local sections of a sheaf of the form~\eqref{Equation : Locally sheaf structure}. When $q = 0$, the coincidence in a \textit{short filtration} can be used to define \textit{local branch numbers} as a heat map of each binary image. 
    \item In the paper, we propose an approximation of local sections in a PH by its PD.  The result allows practitioners to estimate the representatives of local objects in PH.
\end{itemize}
Organization of the paper is as followings.  We present our main results in Section~\ref{sec: Cellular Sheaves Modeling}.  The demonstration of generating local branch numbers of binary images is shown in Section~\ref{sec : Demonstration}. The discussions, future works, and the conclusion are in Section~\ref{Sec : Conclusion}.



\begin{figure}
\begin{center}
 \includegraphics[width=0.40\linewidth]{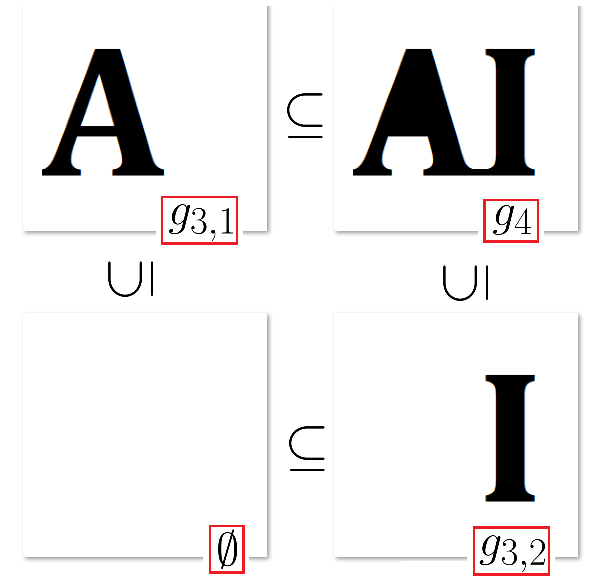}
\end{center}
   \caption{An example of inclusions separated from a filtration of binary images (cf. Figure \ref{FIG : Example of image filtration} (a) $\sim$ (d)). This inclusion relations lead the simplest cellular sheaf structure as in Equation~\eqref{Equation : Locally sheaf structure}).}
\label{fig:Simplest inclusion diagram}
\end{figure}
\section{Cellular Sheaves Modeling}
\label{sec: Cellular Sheaves Modeling}
This section is separated into three parts. In Section~\ref{Subsec: Cellular Sheaves and Coincidence in PH}, we briefly introduce the formal definition of \textit{cellular sheaves over posets}. To define the local sections on cellular sheaves, the \textit{Alexandrov Topology} and the $\mathfrak{B}$-sheaf are also mentioned.  In order to be self-contained, we provide necessary notations and definitions.  For further details on these topics, we refer readers to ~\cite{Curry2015, Ale37, Ale47} and~\cite{QingLiu}. Theorem \ref{Theorem : barcode (2,3) thm} in Section~\ref{Sec : Approximation for Local Sections} approximates the local sections via barcodes.  Finally, in Section~\ref{Sheaves on Images}, we apply our results to analyze local information of binary images.


\subsection{Cellular Sheaves and Coincidence in PH}
\label{Subsec: Cellular Sheaves and Coincidence in PH}
A \textit{poset} (or partially ordered set) $(P,\leq)$ is a non-empty set $P$ equipped with a relation $\leq$ on $P$ which satisfies the following properties: Whenever $x, y, z \in P$, (1) $x \leq x$ (2) $x \leq y$ and $y \leq z$ implies $x \leq z$, and (3) $x \leq y$ and $y \leq x$ implies $x = y$. 
\begin{exam.}
\label{Example : Two main examples of partial orders}
(1) Suppose $X$ is a non-empty set and $2^X$ denotes its power set. Then the inclusion relation $\subseteq$ on $2^X$ is a partial order on $2^X$. In this paper, we mainly consider $\subseteq$ on $2^S$, where $S$ is a subset of $\mathbb{Z}^2$.

\noindent (2) Let $(x_1, x_2, ..., x_n), (y_1, y_2, ..., y_n) \in \mathbb{Z}^n$ be $n$-tuples over $\mathbb{Z}$. Define $(x_1, x_2, ..., x_n) \leq (y_1, y_2, ..., y_n)$ if and only if $x_i \leq y_i$ for all $i \in \{ 1,2, ..., n \}$.  Then $(\mathbb{Z}^n, \leq)$ is a poset. 
\end{exam.}
 

A \textit{cellular sheaf} over a poset $(P,\leq)$ is a rule which assigns each element in $P$ to a vector space over a fixed field $\mathbb{F}$ (\textit{e.g.} $\mathbb{Z}_2$ or $\mathbb{R}$) and preserves the ordering of elements in $P$ via linear transformations. Here is the formal definition:

\begin{def.}
[\cite{robinson2014topological, justinCurryPHDThesis}]
Let $(P,\leq)$ be a poset. A \textbf{celluar sheaf of vector spaces} (over a fixed field $\mathbb{F}$) on $P$ is a rule $\shF$ which consists of the following data:
\begin{itemize}
    \item For each $p \in P$, $\shF$ associates a vector space $\shF(p)$ over $\mathbb{F}$.
    \item For $p \leq q$ in $P$, there is an $\mathbb{F}$-linear transformation $\rho_{p,q} : \shF(p) \rightarrow \shF(q)$ such that $\rho_{p,p}$ is the identity map on $\shF(p)$ and $\rho_{q,r} \circ \rho_{p,q} = \rho_{p,r}$ for every $p \leq q \leq r$. In other words, $\rho_{p,p} = {\rm id}_{\shF(p)}$ and the diagram
    \begin{equation}
    \xymatrix{
    \shF(p) \ar[r]^{\rho_{p,q}} \ar[dr]_{\rho_{p,r}} & \shF(q) \ar[d]^{\rho_{q,r}} \\
    & \shF(r)
    }
    \end{equation}
    is commutative for every $p \leq q \leq r$. 
\end{itemize}
\end{def.}

In other words, a cellular sheaf of vector spaces on a poset $(P,\leq)$ is a functor from the category of elements and partial relations in $P$ to the category of vector spaces over certain field~\cite{robinson2014topological, justinCurryPHDThesis, Curry2015}. By considering the \textit{Alexandrov topology}~\cite{Ale37, Ale47} $\mathfrak{A}$ on the poset $P$ generated by sets $U_p$'s of forms
\begin{equation}
    U_p = \{ q \in P : p \leq q \}
\end{equation}
as an open basis $\mathfrak{B}$ for the topology $\mathfrak{A}$, the assignment $U_p \mapsto \shF(U_p) := \shF(p)$ forms a $\mathfrak{B}$-sheaf structure on the topological space $(P, \mathfrak{A})$~\cite{QingLiu}. The $\mathfrak{B}$-sheaf can be extended as a sheaf of vector spaces over $(P, \mathfrak{A})$, which is the real meaning of ``sheaves on topological spaces''---from the aspect of traditional sheaf theory~\cite{bredon1997sheaf} or algebraic geometry~\cite{Hartshorne, QingLiu}. 

Because $\mathfrak{B} = \{ U_p : p \in P \}$ is an open basis for $\mathfrak{A}$, every open subset $U$ in $(P, \mathfrak{A})$ can be expressed by $U = \bigcup_{p \in U} U_p$. To extend $\shF$ as a sheaf on $(P, \mathfrak{A})$, the $\shF(U)$ is defined as the vector space
\begin{equation}
\label{Eq : The Gluing Technique, inverse limit}
    \shF(U) = \left \{ (s_p)_{p \in U} : s_q = \rho_{p,q}(s_p) \ \forall \ p \leq q \right \},
\end{equation}
where every $(s_p)_{p \in U}$ denotes an element in the Cartesian product $\prod_{p \in U} \shF(p)$ of vector spaces~\cite{ghrist2020cellular, RobinsonNyquist,RobinsonHunting, Curry2015}. Elements in $\shF(U)$ are called \textit{local sections} on $U$, which are the most important targets for observing the coincidence of elements from different $\shF(p)$'s. Here is an example:

\begin{exam.}
\label{Example : Standard Rectangle Example}
Let $P = \{ (0,0), (0,1), (1,0), (1,1) \}$ be a set of tuples in $\mathbb{Z}^2$. Then $(P, \leq)$ is a poset, where $\leq$ is defined as in Example \ref{Example : Two main examples of partial orders} (2). Any cellular sheaf $\shF$ on $P$ can be represented as the commutative diagram
\begin{equation}
\label{Equation : Basic Sheaf Structure}
    \xymatrix@+1.0em{
    \shF(0,1) \ar[r]^{\rho_{(0,1),(1,1)}} & \shF(1,1) \\
    \shF(0,0) \ar[u]^{\rho_{(0,0),(0,1)}} \ar[r]^{\rho_{(0,0),(1,0)}} & \shF(1,0) \ar[u]_{\rho_{(1,0),(1,1)}}
    }
\end{equation}
of vector spaces and linear transformations.

The local sections on $U := U_{(0,1)} \cup U_{(1,0)} = \{ (0,1), (1,0), (1,1) \}$ are pairs $(s,t)$ in $\shF(0,1) \times \shF(1,0)$ such that $\rho_{(0,1),(1,1)}(s) = \rho_{(1,0),(1,1)}(t)$ in $\shF(1,1)$.
\end{exam.}


The concept of local sections described in Example \ref{Example : Standard Rectangle Example} provides a bridge to investigate the relation between $\shF(0,1)$ and $\shF(1,0)$. A local section records elements $s, t$ in $\shF(0,1), \shF(1,0)$ respectively, which would be identified as the same element in $\shF(1,1)$ via $\rho_{\bullet,\bullet}$'s. More concrete examples for digital images and their sheaf structures are discussed in Section~\ref{Sheaves on Images}.

The cellular sheaves in~\eqref{Equation : Basic Sheaf Structure} is a general form of 2-parameter persistent homology over the poset $P$. In the paper, we mainly consider one-parameter PH
\begin{equation}
\label{Eq : 1-parameter PH}
    H_{q}(X_0) \xrightarrow{\rho_{0,1}} H_{q}(X_1) \xrightarrow{\rho_{1,2}} \cdots \xrightarrow{} H_{q}(X_n)
\end{equation}
of a filtration $\emptyset = X_0 \subseteq X_1 \subseteq \cdots \subseteq X_n$ of topological spaces\footnote{In the paper, we consider $H_{q}(X_i)$'s and $\rho_{i,i+1}$'s as vector spaces and linear transformations over the binary field $\mathbb{Z}_2$}, where $\rho_{i,i+1}$'s appear in \eqref{Eq : 1-parameter PH} are linear transformations induced by the inclusions $X_{i} \hookrightarrow X_{i+1}$~\cite{Greenberg}. For integers $0 \leq i < j \leq n$, 
 is defined as
\begin{equation}
    \rho_{i,j} = \rho_{j-1,j} \circ \cdots \circ \rho_{i,i+1}
\end{equation}
and $\rho_{i,i}$ as the identity map of $H_q(X_i)$.

Next we define the \textit{coincidence of pair} $(s_i,s_j) \in H_q(X_i) \oplus H_q(X_j)$\footnote{For finitely many homologies, we use the notation $\oplus$ to replace $\times$.} to $H_q(X_k)$ with $i \leq j \leq k$ in~\eqref{Eq : 1-parameter PH}.  The motivation of the definition is to capture the merging relation between certain $s_i$ and $s_j$ in the filtration which can be viewed as a local version of the definition of barcodes (cf. Lemma \ref{App Lemma}).
\begin{def.}
Let $\emptyset = X_0 \subseteq X_1 \subseteq X_2 \subseteq \cdots \subseteq X_n$ be a filtration of topological spaces and $0 \rightarrow H_{q}(X_1) \xrightarrow{} H_{q}(X_2) \xrightarrow{} \cdots \xrightarrow{} H_{q}(X_n)$ be its PH. For $q \geq 0$, $i, j \in \{ 1,2, ..., n \}$, $s_i \in H_{q}(X_i)$, $s_j \in H_{q}(X_j)$, and $\max\{ i,j \} \leq k \leq n$, we say $s_i$ and $s_j$ \textbf{coincide} at $k$ if $\rho_{i,k}(s_i) = \rho_{j,k}(s_j)$.
\end{def.}

In other words, if we extract $H_q(X_i)$, $H_q(X_j)$, and $H_q(X_k)$ from the PH, and connect them by the natural linear transformations, then the coincidence of pairs $(s_i,s_j)$ can be regarded as the local sections of in $H_q(X_i) \oplus H_q(X_j)$ with respect to the sheaf of form~\eqref{Equation : Basic Sheaf Structure}. This can be formulated as the following theorem.
\begin{theorem}
\label{Thm : Say coincide as a local section}
Let $\emptyset = X_0 \subseteq X_1 \subseteq X_2 \subseteq \cdots \subseteq X_n$ be a filtration of topological spaces, $q \geq 0$, $s_i \in H_{q}(X_i)$, and $s_j \in H_{q}(X_j)$. Then $s_i$ and $s_j$ coincide at $k \geq \max\{ i,j \}$ if and only if $(s_i,s_j) \in H_q({X_{i}}) \oplus H_q({X_{j}})$ is a local section with respect to the cellular sheaf~\eqref{Equation : Locally sheaf structure}.
\begin{equation}
\label{Equation : Locally sheaf structure}
    \xymatrix@+0.0em{
    H_q(X_{i}) \ar[r]^{\rho_{i,k}} & H_q(X_k) \\
    & H_q(X_{j}) \ar[u]_{\rho_{j,k}}
    }
\end{equation}
\end{theorem}
\begin{proof}
By the same arguments as in Example \ref{Example : Standard Rectangle Example}, an element  $(s_i,s_j) \in H_q({X_{i}}) \oplus H_q({X_{j}})$ is a local section of the cellular sheaf~\eqref{Equation : Locally sheaf structure} if and only if $\rho_{i,k}(s_i) = \rho_{j,k}(s_j)$, as desired.
\end{proof}


For example, in the filtration (a) $\sim$ (d) in Figure~\ref{FIG : Example of image filtration}, the component ``I" has barcode $(3,4)$ since it was born at $g_3$ and finally be merged into ``A". However, the persistence diagram (Figure~\ref{FIG : Example of image filtration} (e)) only shows that ``I" has barcode $(3,4)$ while the merging information of ``I" and ``A" could not be concluded in the diagram.

As in Theorem~\ref{Thm : Say coincide as a local section}, this local behavior between ``A" and ``I" is encoded as a local section of the cellular sheaf~\eqref{Equation : Simplest Sheaf diagram in homology}, by separating image $g_3$ into two parts $g_{3,1} = \rm ``A"$ and $g_{3,2} = \rm ``I"$. The inclusions $g_{3,1}^{-1}(0) \hookrightarrow g_4^{-1}(0)$ and $g_{3,2}^{-1}(0) \hookrightarrow g_4^{-1}(0)$ in Figure~\ref{fig:Simplest inclusion diagram} lead the diagram
\begin{equation}
\label{Equation : Simplest Sheaf diagram in homology}
    \xymatrix@+0.0em{
    H_0(g_{3,1}^{-1}(0)) \ar[r]^{\rho_1} & H_0(g_{4}^{-1}(0)) \\
                                         & H_0(g_{3,2}^{-1}(0)) \ar[u]_{\rho_2}
    }
\end{equation}
of $0$-dimensional homologies where $\rho_1$ and $\rho_2$ are linear transformations induced by inclusions. Note that
\begin{equation}
\begin{split}
    H_0(g_{3,1}^{-1}(0)) &= {\rm span}_{\mathbb{Z}_2} \{ {\rm ``A"} \},\\  H_0(g_{3,2}^{-1}(0)) &= {\rm span}_{\mathbb{Z}_2} \{ {\rm ``I"} \}
\end{split}
\end{equation}
and $\rho_1({\rm ``A"}) = \rho_2({\rm ``I"})$ in $H_0(g_{4}^{-1}(0))$. That is, the pair $({\rm ``A", ``I"})$ is a local section which records that ``A" and ``I" finally merged into the unique connected component of $g_4$.

\subsection{Approximation for Local Sections}
\label{Sec : Approximation for Local Sections}
Although elements in homology $H_{\bullet}(X)$ can be defined precisely (even discretely with $\mathbb{Z}_2$ as the field of coefficients), it is still difficult to represent elements in $H_{\bullet}(X)$ by comparing all representatives. To tackle this, we propose a method that uses barcodes of PH to approximate the local sections.

Here we recall the definition of barcodes of elements in PH~\cite{EdelsbrunnerHarerbook2010}: let $\emptyset = X_0 \subseteq X_1 \subseteq \cdots \subseteq X_n$ be a filtration of topological spaces and $\{ 0 \} \rightarrow H_q(X_1) \rightarrow \cdots \rightarrow H_q(X_n)$ be its PH, an element $s_i \in H_q(X_i)$ ($i \geq 1$) is said to have \textit{barcodes} $(i,j)$ with $i < j$ if it satisfies the following two properties:
\begin{itemize}
    \item $s_i \notin {\rm im}(\rho_{i-1,i})$;
    \item $\rho_{i,j-1}(s_i) \notin {\rm im}(\rho_{i-1,j-1})$ and $\rho_{i,j}(s_i) \in {\rm im}(\rho_{i-1,j})$,
\end{itemize}
where the death $j$ may be $\infty$ if $s_i$ is still alive at $n$.   

As Edelsbrunner and Harer mentioned in~\cite{EdelsbrunnerHarerbook2010}, the $s_i$ has death $j$ if it merges with an older class in $H_q(X_{j-1})$. The following approximation lemma can be viewed as a local version of barcodes of elements, which approximates the upper bound of deaths for any $s_i, s_j$ who are coinciding at some $k$.

\begin{lemma}
[Approximation Lemma]
\label{App Lemma}
Let $\emptyset = X_0 \subseteq X_1 \subseteq \cdots \subseteq X_n$ be a filtration of topological spaces, $q \geq 0$, $0 \leq i < j$, $s_i \in H_{q}(X_i)$, and $s_j \in H_{q}(X_j)$ which has barcode $(j,d)$. If $s_i, s_j$ coincide at $k \geq j$, then $d \leq k$.
\end{lemma}
\begin{proof}
We first note that $k$ must be strictly larger than $j$: If $k = j$, then $s_j = \rho_{j,j}(s_j) = \rho_{i,j}(s_i) = (\rho_{j-1,j} \circ \rho_{i,j-1})(s_i)$, this shows that $s_j \in {\rm im}(\rho_{j-1,j})$ and contradicts to $s_j$ is born at $j$. Hence we have $k > j$.

Suppose $d > k$, then $i < j < k \leq d-1$. In particular, $s_j$ doesn't die at $k$. By definition of death, either $\rho_{j,k-1}(s_j) \in {\rm im}(\rho_{j-1, k-1})$ or $\rho_{j,k}(s_j) \notin {\rm im}(\rho_{j-1, k})$. Because
\begin{equation*}
    \rho_{j,k}(s_j) = \rho_{i,k}(s_i) = \rho_{j-1,k}(\rho_{i,j-1}(s_i)) \in {\rm im}(\rho_{j-1,k}),
\end{equation*}
we must have $\rho_{j,k-1}(s_j) \in {\rm im}(\rho_{j-1, k-1})$. Because $j \leq k - 1 \leq d - 1$, we have
\begin{equation}
\begin{split}
    \rho_{j,d-1}(s_j) &= (\rho_{k-1,d-1} \circ \rho_{j,k-1})(s_j) \in {\rm im}(\rho_{j-1,d-1})
\end{split}
\end{equation}
since $\rho_{k-1,d-1}({\rm im}(\rho_{j-1, k-1})) = {\rm im}(\rho_{j-1,d-1})$. However, $s_j \notin {\rm im}(\rho_{j-1,d-1})$ since $s_j$ dies at $d$, a contradiction. This shows that $d \leq k$.
\end{proof}
The approximation lemma can be used for estimating whether certain pair $(s_i,s_j)$  coincides at some $k \geq \max\{ i, j \}$ or not. To adapt the lemma, we say a filtration which is \textit{short} if it has the form $\emptyset = X_0 \subseteq X_1 \subseteq X_2 \subseteq X_3$.
\begin{theorem}
\label{Theorem : barcode (2,3) thm}
Let $\mathcal{F} : \emptyset = X_0 \subseteq X_1 \subseteq \cdots \subseteq X_n$ be a filtration of topological spaces, $q \geq 0$, $0 \leq i < j < k$, and $s_j \in H_{q}(X_j)$ which is born at $j$. Define short filtration
\begin{equation}
\mathcal{G} : \emptyset = Y_0 \subseteq Y_1 \subseteq Y_2 \subseteq Y_3
\end{equation}
where $Y_0 = X_0$, $Y_1 = X_i$, $Y_2 = X_j$, and $Y_3 = X_k$, then $s_j$ is coincide to some $s_i \in H_{q}(X_i)$ at $k$ in $\mathcal{F}$ if and only if $s_j \in H_q(Y_2)$ has barcode $(2,3)$ in $\mathcal{G}$.
\end{theorem}
\begin{proof}
We first assume there is an $s_i \in H_q(X_i)$ such that $s_i$ and $s_j$ coincide at $k$ in $\mathcal{F}$. Because $s_j$ is born at $j$ in $\mathcal{F}$, $\rho_{i,j}(s_i) \neq s_j$. Hence $s_j$ is also born at $2$ in $\mathcal{G}$. In $\mathcal{G}$, $s_i$ and $s_j$ also coincide at $3$, by the approximation lemma, the death of $s_j$ in must $\leq 3$, and it forces that $s_j$ has barcode $(2,3)$ in $\mathcal{G}$.

Conversely, suppose $s_j \in H_q(Y_2)$ has barcode $(2,3)$ in $\mathcal{G}$, then $\rho^{\mathcal{G}}_{2,3}(s_j) \in {\rm im}(\rho^{\mathcal{G}}_{1,3})$ where $\rho^{\mathcal{G}}_{\bullet,\bullet}$'s are the induced linear transformations from $\rho_{\bullet,\bullet}$ in $\mathcal{F}$. This shows that there is an $s_i \in H_q(Y_1) = H_q(X_i)$ such that $\rho_{i,k}(s_i) = \rho^{\mathcal{G}}_{1,3}(s_i) = \rho^{\mathcal{G}}_{2,3}(s_j) = \rho_{j,k}(s_j)$ i.e., $s_i$ and $s_j$ coincide at $k$ in $\mathcal{F}$.
\end{proof}
\subsection{Sheaf Structures on Binary Images}
\label{Sheaves on Images}
The subsection is separated into two parts. We first introduce some notations and terminologies for binary images, and then introduce the \textit{short filtrations} for single binary image and apply its sheaves to compute the local branch numbers. 
\subsubsection*{Binary Images}
A 2-dimensional \textit{digital image} can be regarded as a function $f : S \rightarrow \mathbb{R}_{\geq 0}$, where $S$ is a subset of the 2-dimensional grid $\mathbb{Z}^2$. Typically, $S = ([a,b] \times [c,d]) \cap \mathbb{Z}^2$ is a rectangle, called the set of pixels of image $f$~\cite{serra1984image, Soille2003, Najman-Mathematical-Morphology}. A digital image is said to be \textit{binary} if it has range $\{ 0, 1 \}$ where $0$ and $1$ usually denote the pixel values of black and white respectively. The set of all black pixels in $S$ is the pre-image set $f^{-1}(0)$. 

Suppose $f_1, ..., f_n$ is a sequence of binary images on $S$ which satisfies 
\begin{equation}
\label{Equation : Filtration of binary images}
    f_1^{-1}(0) \subseteq f_2^{-1}(0) \subseteq \cdots \subseteq f_n^{-1}(0),
\end{equation}
then~\eqref{Equation : Filtration of binary images} is called a \textit{filtration} of binary images (\textit{e.g.} Figure~\ref{FIG : Example of image filtration} (a) $\sim$ (d)). 

By viewing every set of black pixels $f^{-1}(0) \subseteq \mathbb{Z}^2$ of a binary image $f : S \rightarrow \{ 0, 1\}$ as a collection of squares in $\mathbb{R}^2$ i.e., a \textit{cubical complex} embedded in $\mathbb{R}^2$~\cite{Wagner2012,kaczynski2004computational, gudhi:FilteredComplexes, gudhi:urm}. Because the union of squares can be viewed as a subspace of $\mathbb{R}^2$, the computation of homologies are valid. 

For a $2$-dimensional image $f$, the homology of $f^{-1}(0)$ detects the numbers $(\beta_0, \beta_1)$ of connected components and $1$-dimensioanl holes form by black pixels of $f$ respectively, which are called the \textit{Betti numbers}~\cite{Greenberg, EdelsbrunnerHarerbook2010}. More precisely, $\beta_q = \dim_{\mathbb{Z}_2} H_q(f^{-1}(0))$, $q = 0,1$. For example, as (c) in Figure \ref{FIG : Example of image filtration}, if $f = g_3$ is the image ``AI", then 
\begin{equation}
\begin{split}
H_0(f^{-1}(0)) &= {\rm span}_{\mathbb{Z}_2} \{ {\rm ``A"}, {\rm ``I"} \}, \\
H_1(f^{-1}(0)) &= {\rm span}_{\mathbb{Z}_2} \{ {\rm ``A"} \}.    
\end{split}
\end{equation}
because ``A" and ``I" represent different connected components of ``AI" and ``A" contains a 1-dimensional hole. In particular, $\beta_0 = 2$ and $\beta_1 = 1$.


\subsubsection*{Cellular Sheaves on Single Binary Image}
We propose a method for analyzing connecting relations between local objects in a binary image by using sheaf structure. Let $X$ be a topological space and $X_1, X_2$ be subspaces of $X$. We imitate the idea in Theorem \ref{Thm : Say coincide as a local section} to construct the natural cellular sheaf structure
\begin{equation}
\label{Equation : cellular sheaf of single space}
    \xymatrix@+0.5em{
    H_q(X_{1}) \ar[r]^{\rho_1} & H_q(X) \\
    & H_q(X_{2}) \ar[u]_{\rho_2}
    }
\end{equation}
where $\rho_1, \rho_2$ are induced by the inclusion maps. Now we have two short filtrations
\begin{equation}
    \begin{split}
        \mathcal{G}_1 &: \emptyset \subseteq X_1 \subseteq X_1 \cup X_2 \subseteq X, \\
        \mathcal{G}_2 &: \emptyset \subseteq X_2 \subseteq X_1 \cup X_2 \subseteq X.
    \end{split}
\end{equation}
and persistent homologies
\begin{equation}
\label{Equation : Two short persistent homologies}
    \begin{split}
        & 0 \xrightarrow{} H_q(X_1) \xrightarrow{\omega_1} H_q(X_1 \cup X_2) \xrightarrow{\gamma} H_q(X), \\
        & 0 \xrightarrow{} H_q(X_2) \xrightarrow{\omega_2} H_q(X_1 \cup X_2) \xrightarrow{\gamma} H_q(X),
    \end{split}
\end{equation}
whenever $q \geq 0$, where $\omega_1, \omega_2$ and $\gamma$ are induced by the inclusion maps and $\gamma \circ \omega_i = \rho_i$ for $i = 1,2$.

\begin{remark}
We note that although short filtrations $\mathcal{G}_1$ and $\mathcal{G}_2$ share the same cellular sheaf structures~\eqref{Equation : cellular sheaf of single space}, the persistence diagrams of $\mathcal{G}_1$ and $\mathcal{G}_2$ are different. For example, if $X = \mathbb{R}^2$, $X_1 = S^1 := \{ (x,y) : x^2 + y^2 = 1 \}$ and $X_2 = B_{(0,0)}(1) := \{ (x,y) : x^2 + y^2 < 1 \}$ are the unit circle and unit open disk in $\mathbb{R}^2$, then $\mathcal{P}_1(\mathcal{G}_2)$ is empty while $\mathcal{P}_1(\mathcal{G}_1)$ has barcode $(1,2)$.
\end{remark}

In addition, when ${\rm cl}_X(X_1)$ and ${\rm cl}_X(X_2)$ are disjoint\footnote{If $X$ is a topological space and $A$ is a subset of $X$, ${\rm cl}_X(A)$ denotes the closure of $A$ in $X$.}, the canonical mapping
\begin{equation}
    \omega_1 + \omega_2 : H_q(X_1) \oplus H_q(X_2) \rightarrow H_q(X_1 \cup X_2)
\end{equation}
defined by $(\xi_1, \xi_2) \longmapsto \xi_1 + \xi_2$ is an isomorphism (\cite{Greenberg} Proposition (9.5)). In this case, $\omega_1$ and $\omega_2$ in~\eqref{Equation : Two short persistent homologies} are one-to-one. Because $\omega_1 + \omega_2$ is bijective, every element $s$ in $H_q(X_1 \cup X_2)$ can be uniquely represented by $s_1 + s_2$, where $s_i = \omega_i(\widetilde{s_i})$ for some $\widetilde{s_i} \in H_q(X_i)$, $i = 1,2$. In particular, we have the following theorem. 

\begin{figure}
\begin{center}
\subfloat[$X$]{\includegraphics[scale=0.4]{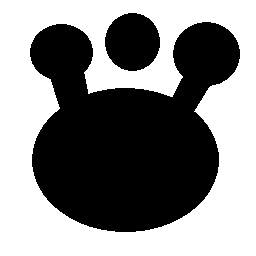}}~
\subfloat[$X_1$]{\includegraphics[scale=0.4]{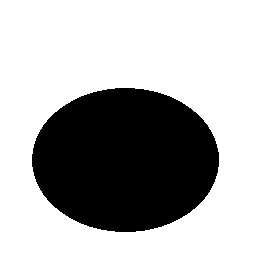}}~
\subfloat[$X_2$]{\includegraphics[scale=0.4]{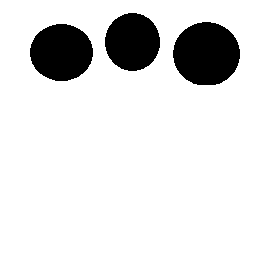}}~
\subfloat[$X_1 \cup X_2$]{\includegraphics[scale=0.40]{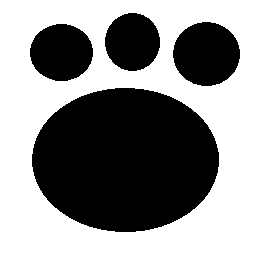}}
\end{center}
\centering
\caption{An example of spaces $X$ in $\mathbb{R}^2$ and its subspaces $X_1, X_2$, which satisfy ${\rm cl}_X(X_1) \cap {\rm cl}_X(X_2) = \emptyset$.}
\label{FIG : Example of Branch Numbers}
\end{figure}

\begin{theorem}
\label{Thm for single binary image}
Let $X, X_1, X_2$, $q \geq 0$ and $\rho_i, \omega_i, \gamma$, $i = 1,2$ be defined as above. If ${\rm cl}_X(X_1) \cap {\rm cl}_X(X_2) = \emptyset$, then the following hold:
\begin{itemize}
    \item[$\rm (a)$] The persistence diagram $\mathcal{P}_q(\mathcal{G}_1)$ has no barcodes of birth $= 2$ if and only if $H_q(X_2) = \{ 0 \}$.
    \item[$\rm (b)$] For $s_2 = \omega_2(\widetilde{s_2}) \in H_q(X_1 \cup X_2)$, $s_2 \neq 0$ if and only if it is born at $2$ in $\mathcal{G}_1$.
    \item[$\rm (c)$] For non-zero $\widetilde{s_2} \in H_q(X_2)$, $(\widetilde{s_1}, \widetilde{s_2}) \in H_q(X_1) \oplus H_q(X_2)$ is a local section in~\eqref{Equation : cellular sheaf of single space} for some $\widetilde{s_1} \in H_q(X_1)$ if and only if $s_2 := \omega_2(\widetilde{s_2})$ has barcode $(2,3)$ in $\mathcal{G}_1$.
    \end{itemize}
\end{theorem}
\begin{proof}
(a) Because $\omega_1 +\omega_2$ is an isomorphism, $P_q(\mathcal{G}_1)$ has not barcodes of birth $= 2$ if and only if ${\rm im}(\omega_1) = H_q(X_1 \cup X_2) = {\rm im}(\omega_1 +\omega_2)$ if and only if $\omega_2$ is the zero map if and only if $H_q(X_2) = \{ 0 \}$ (since $w_2$ is one-to-one). 

(b) The converse direction is trivial. If $s_2 = \omega_2(\widetilde{s_2}) = (\omega_1 + \omega_2)(0,\widetilde{s_2})$ is non-zero, then $s_2 \notin {\rm im}(\omega_1) = {\rm im}(\omega_1 +\omega_2)$ since $\omega_1 +\omega_2$ is one-to-one.

(c) By (b), $s_2$ is born at 2 since $\widetilde{s_2} \neq 0$ and $\omega_2$ is one-to-one. To adapt Theorem \ref{Theorem : barcode (2,3) thm}, it is sufficient to note that $\rho_1(\widetilde{s_1}) = s_2$ if and only if $\gamma(\omega_1(\widetilde{s_1})) = s_2$ i.e., $\widetilde{s_1}$ and $s_2$  coincide at $3$ in $\mathcal{G}_1$. Now (c) follows from Theorem \ref{Theorem : barcode (2,3) thm} immediately.
\end{proof}  
As in Figure \ref{FIG : Example of Branch Numbers}, the $0$-dimensional PD $\mathcal{P}_0(\mathcal{G}_1)$ of $\shG_1$ is the multiset $\{ (1,+\infty), (2,3), (2,3) \}$ while $\mathcal{P}_0(\mathcal{G}_2) = \{ (1,+\infty), (2,3) \}$. It indicates that the space $X_1$ has two branchs in $X$. From the $X_2$'s point of view, it has single branch since $X_1$ is the unique component in $X$ which connects to $X_2$. Now the following definitions can be established:

\begin{def.}
Let $X, X_1, X_2$, and $\mathcal{G}_1, \mathcal{G}_2$, $i = 1,2$ be defined as in Theorem \ref{Thm for single binary image}, then the \textbf{local branch number} of $X_1$ with respect to $X_2$ in $X$ is defined as the number of barcodes $(2,3)$ in the multiset $\mathcal{P}_0(\mathcal{G}_1)$, denoted by $b_0(X_1;X_2)$. 
\end{def.}

\begin{def.}
A \textbf{system of patches} of a topological space $X$ is a (finite) collection of pairs $(X_{1}^{(i)}, X_{2}^{(i)})$'s of subspaces of $X$ such that ${\rm cl}_X(X_1^{(i)}) \cap {\rm cl}_X(X_2^{(i)}) = \emptyset$ for every $i$. 
\end{def.}

\section{Demonstration}
\label{sec : Demonstration}
In the section, we demonstrate how do we use the proposed method to automatically detect local branch numbers of a binary image. The demonstration aims to automatically generated a heat map $b_0(f)$ for representing local branch numbers of local regions of a binary image $f : S \rightarrow \{ 0,1 \}$ where $S = ([a,b] \times [c,d]) \cap \mathbb{Z}^2$. Clearly, the $b_0(f)$ depends on how do we define patches $\{ (X_1^{(i)}, X_2^{(i)}) \}_{i = 1}^n$ of $f^{-1}(0)$, hence $b_0(f)$ is not unique. 

To construct a system of patches of a binary image, for each local window $S' := ([a',b'] \times [c',d']) \cap \mathbb{Z}^2$ with $a \leq a' \leq b' \leq b$ and $c \leq c' \leq d' \leq d$, we define $\widehat{X_1} = f^{-1}(0) \cap S'$ and $X_2 = f^{-1}(0) \setminus \widehat{X_1}$. Next, we define
\begin{equation}
    X_1 = \widehat{X_1} \setminus \{ (x,y) \in \widehat{X_1} : x = a' \ {\rm or} \ y = b' \}.
\end{equation}
In other words, we obtain $X_1$ by removing all black pixels which belong to the boundary of $S'$. Because finite cubic complexes in $\mathbb{R}^2$ are finite unions of closed $\mathbb{R}^2$-squares, they are closed in $\mathbb{R}^2$ (and so do $X$). Hence we must have ${\rm cl}_X(X_1) \cap {\rm cl}_X(X_2) = \emptyset$. 

\begin{figure}
\begin{center}
\subfloat[Input image]{\includegraphics[scale=0.2]{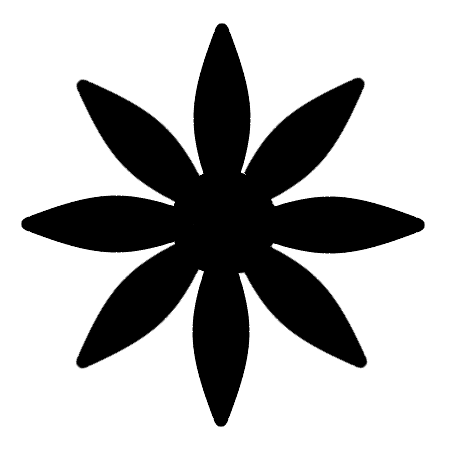}}~~~~~
\subfloat[Windows]{\includegraphics[scale=0.2]{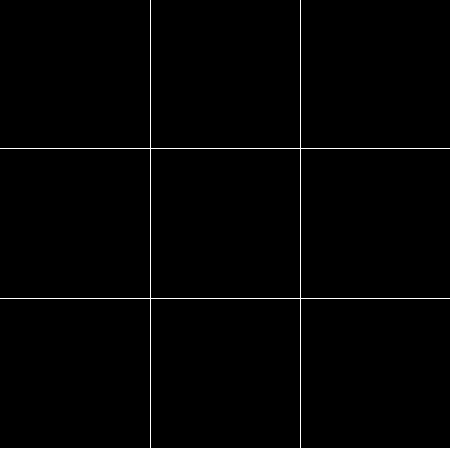}}~~~~~
\subfloat[$\{ X_1^{(i)} \}_{i = 1}^9$]{\includegraphics[scale=0.2]{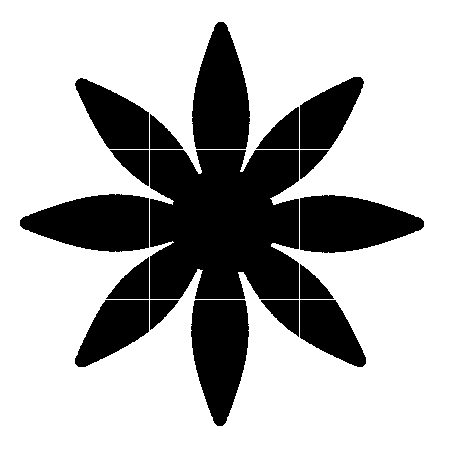}}~~~~~
\subfloat[Branch numbers]{\includegraphics[scale=0.2]{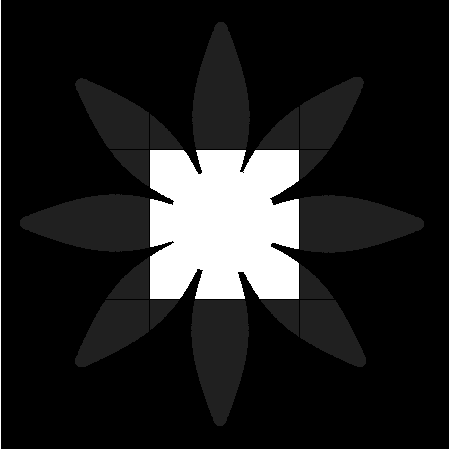}}
\end{center}
\centering
\caption{An example of windows, patches and branch numbers.}
\label{FIG : Image, Patches, and Heatmap}
\end{figure}

For example, as in Figure~\ref{FIG : Image, Patches, and Heatmap}, if we consider local windows in (b), then for image (a), the family $\{ X_1^{(i)} \}_{i = 1}^9$ of black pixels in (c) induces a system of patches $\{ (X_1^{(i)}, X_2^{(i)}) \}_{i = 1}^9$, and (d) is the corresponding heat map of local branch numbers. Actually, if we exclude the black pixels, the heat map (d) corresponds to the matrix      
\begin{equation}
    \begin{array}{|c|c|c|}
    \hline
1 & 1 & 1\\  \hline 
1 & 8 & 1\\  \hline 
1 & 1 & 1 \\
\hline
\end{array}
\end{equation}
of local branch numbers, which is compatible with our intuition about branch numbers of local objects. With the same windows in Figure~\ref{FIG : Image, Patches, and Heatmap} (b), we provide more heat maps and corresponding matrices of squared images as examples for explaining our method. Every image was resized into $100 \times 100$ pixels, and its heat map can be generated in nearly 5 seconds\footnote{Local branch numbers are invariant under the transformation of similar objects in $\mathbb{R}^2$.}.     

\begin{figure}
\begin{center}
\subfloat[Input image]{\includegraphics[scale=0.16]{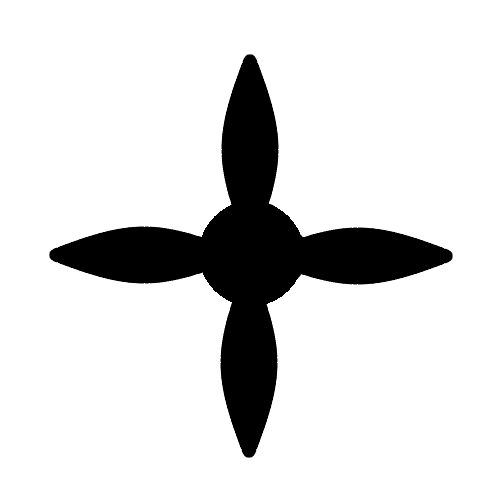}}~
\subfloat[Heat map]{\includegraphics[scale=0.8]{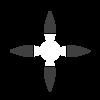}}~
\subfloat[Matrix]{\includegraphics[scale=0.178]{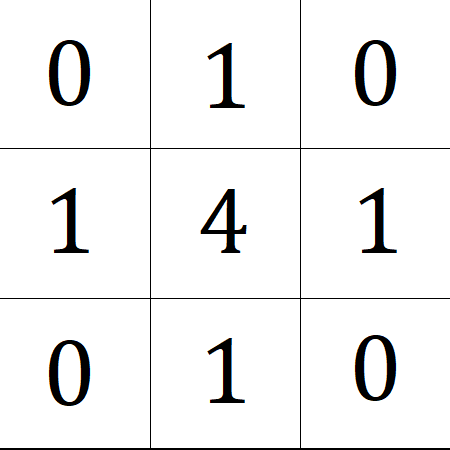}}
\\
\subfloat[Input image]{\includegraphics[scale=0.16]{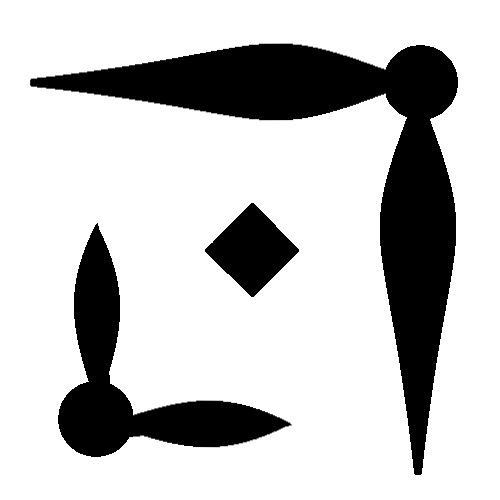}}~
\subfloat[Heat map]{\includegraphics[scale=0.8]{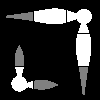}}~
\subfloat[Matrix]{\includegraphics[scale=0.178]{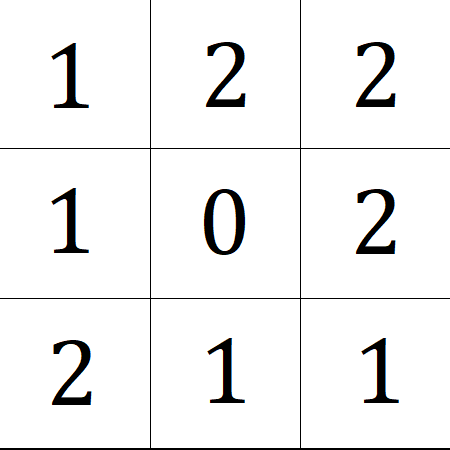}}
\end{center}
\centering
\caption{More examples of images, heat maps, and matrices.}
\label{FIG : More Images, and Heatmaps}
\end{figure}

\begin{figure}
\begin{center}
\subfloat[Input image]{\includegraphics[scale=0.8]{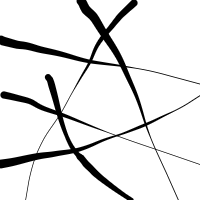}}~
\subfloat[Heat map]{\includegraphics[scale=0.448]{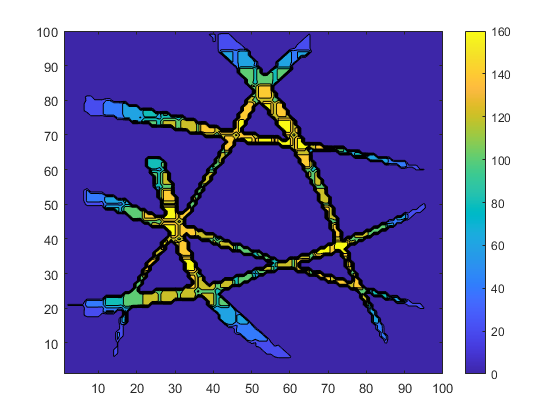}}
\end{center}
\centering
\caption{Image and its heat map of local branch numbers.}
\label{FIG : Sliding Window Approach}
\end{figure}

\begin{figure}
\begin{center}
\subfloat[Selected Images in UIUC]{\includegraphics[scale=0.45]{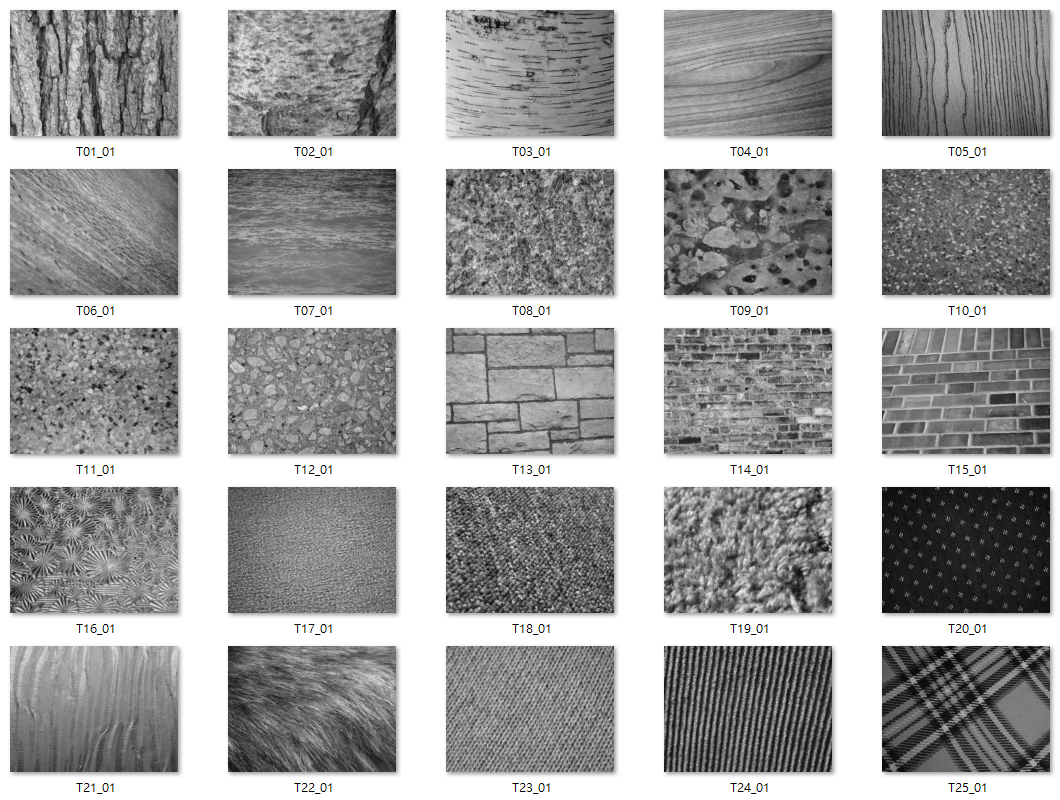}}\\
\subfloat[Heat maps of branch numbers of selected Images ]{\includegraphics[scale=0.48]{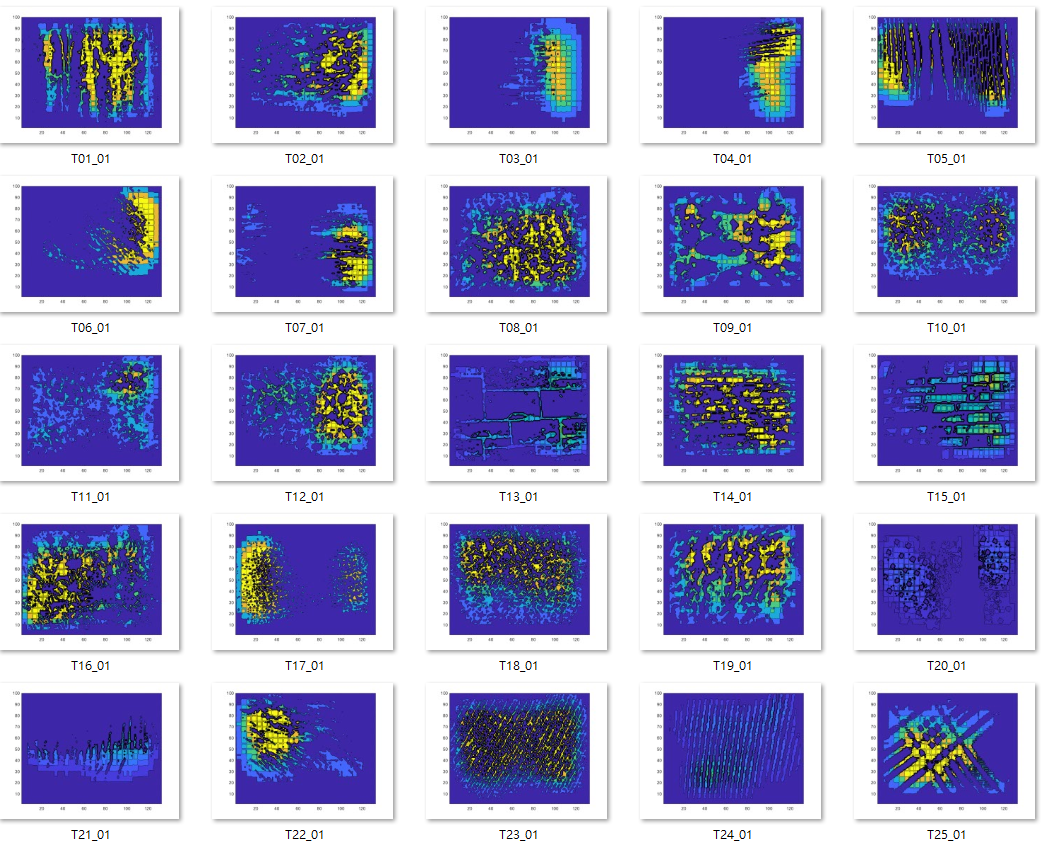}}
\end{center}
\centering
\caption{Images in UIUC dataset and their heat maps.}
\label{FIG : Demonstration on UIUC}
\end{figure}

Alternatively, we can also apply the sliding window algorithm~\cite{Braverman2016} to compute the local branch numbers of local windows. In the demonstration, we unify each binary image $f$ to have size of $100 \times 100$ pixels, and consider three local windows of sizes $10 \times 10$, $20 \times 20$, and $30 \times 30$. For each local window we obtain heat map $b_0^{n}(f)$, $n = 10, 20, 30$, and finally output $b_0(f)$ as the sum of all $b_0^{n}(f)$'s. Figure~\ref{FIG : Sliding Window Approach} and Figure~\ref{FIG : Demonstration on UIUC} are examples.  The codes of the demonstration were run by {\tt Matlab 2020b} based on {\tt Windows 10}, and the PDs were computed by the software {\tt Perseus} \cite{perseus}.

\section{Discussion and Conclusion}
\label{Sec : Conclusion}
Based on our investigation, the proposed work is the first one that combines sheaf structures and persistence diagrams to capture the local branch numbers of binary images. We use the simplest sheaf structure~\eqref{Equation : Locally sheaf structure} to derive the merging relation between pairs of local objects. One future direction is to investigate the meaning and applications of local sections of homologies in $q \geq 1$, and consider more complicated sheaf structures, such as the coincidence of $n$-tuples $(s_1, ..., s_n)$ and their topological insights behind the algebraic coincidence.

Although the form of the diagram~\eqref{Equation : Locally sheaf structure} is the simplest form of the zigzag homology~\cite{CarlssonSinghZomorodian, SCARAMUCCIA2020101623}, multi-parameter persistent homology~\cite{CarlssonSinghZomorodian, SCARAMUCCIA2020101623}, and cellular sheaves~\cite{Curry2015, robinson2014topological, RobinsonNyquist}, the fruitful tools developed in the theories were not considered in the work, such as cellular sheaf cohomology and multi-persistence. Analyzing digital images by combining these advanced tools is also a future direction.

We also demonstrate our method on UIUC dataset~\cite{1453514} in Figure \ref{FIG : Demonstration on UIUC}, which is a well-known dataset in texture classification task~\cite{Kolouri2016}. The UIUC dataset provides 25 classes of images in different textures. We select one image per class and compute its heat map, where we transform each image to a binary one by setting its average pixel value as a threshold. The result shows that images in different classes have different characteristics of heat maps. For example, heat maps of textures with regular patterns may distribute more uniformly, and heat maps of natural textures (\textit{e.g.} wood surfaces) may contain small and non-regular pieces that have high $b_0(f)$. As a low-level feature, the local branch numbers may provide useful information in certain image classification tasks.    

Lastly, since the local branch numbers provide a heat map of an image, it can be viewed as a natural attention map. Local parts who have higher local branch values show that they are the joint areas crossed by different objects, and might be important in some image analysis tasks (\textit{e.g.} medical images). Because every image and its heat map have the same size, the later one can be viewed as an additional channel of an image input in neural network models (\textit{e.g.} AlexNet~\cite{NIPS2012_c399862d} or ResNet~\cite{ResNet7780459}). We also plan to integrate this geometrically explainable feature into current DNN models for some computer vision tasks.

To conclude this paper, we highlight our contribution. We proposed a new theory to generate the local branch numbers in a binary image by using cellular sheaves on short persistent homology. Moreover, the approximation theorem shows that local sections in cellular sheaves can be computed via barcodes in persistence diagrams of short filtrations. Because the local branch numbers are purely provided by the intrinsic geometry of the given image, no learning procedures needed in the method. We also implemented the code for generating local branch numbers, which has the potential to be integrated into DNN models.    

\section*{Acknowledgments}
Chuan-Shen Hu is supported by the projects MOST 108-2119-M-002-031 and MOST 108-2115-M-003-005-MY2 hosted by the Ministry of Science and Technology in Taiwan.
{\small
\bibliographystyle{ieee_fullname}
\bibliography{egbib}
}

\end{document}